\pdfoutput=1
\documentclass[11pt]{article}

\usepackage[final]{acl}

\usepackage{times}
\usepackage{latexsym}

\usepackage[T1]{fontenc}
\usepackage[utf8]{inputenc}
\usepackage{arydshln}
\usepackage{microtype}
\usepackage{amsmath}
\usepackage{makecell}
\usepackage{mathtools}
\usepackage{inconsolata}

\usepackage{graphicx}
\usepackage{booktabs}
\usepackage[textsize=tiny]{todonotes}
\usepackage[normalem]{ulem}
\usepackage{ifthen}
\newboolean{showcomments}
\setboolean{showcomments}{true}  
\usepackage{enumitem}
\usepackage{fontawesome5}
\newcommand{\circled}[1]{\text{\textcircled{\scriptsize #1}}}

\ifthenelse{\boolean{showcomments}}{
\newcommand\li[1]{\textcolor{blue}{Li: #1}}
\newcommand\pw[1]{\textcolor{purple}{PW: #1}}
\newcommand\simon[1]{\textcolor{cyan}{Du: #1}}
}
{
\newcommand\li[1]{}
\newcommand\pw[1]{}
\newcommand\simon[1]{}
}

\title{Exploring How Generative MLLMs Perceive More Than CLIP \\with the Same Vision Encoder}

\author{Siting Li, Pang Wei Koh, Simon Shaolei Du \\
        University of Washington \\ \texttt{\{sitingli,pangwei,ssdu\}@cs.washington.edu}}

\begin{document}
\maketitle
\def\thefootnote{}\footnotetext{\faGithub~~~~\url{https://github.com/lst627/CLIP-Embeds}}\def\thefootnote{\arabic{footnote}}
\begin{abstract}
Recent research has shown that CLIP models struggle with visual reasoning tasks that require grounding compositionality, understanding spatial relationships, or capturing fine-grained details. One natural hypothesis is that the CLIP vision encoder does not embed essential information for these tasks. However, we find that this is not always the case: The encoder gathers query-relevant visual information, while CLIP fails to extract it. In particular, we show that another branch of Vision-Language Models (VLMs), Generative Multimodal Large Language Models (MLLMs), achieve significantly higher accuracy than CLIP in many of these tasks using the \emph{same} vision encoder and weights, indicating that these Generative MLLMs \emph{perceive more}—as they extract and utilize visual information more effectively. We conduct a series of controlled experiments and reveal that their success is attributed to multiple key design choices, including patch tokens, position embeddings, and prompt-based weighting. On the other hand, enhancing the training data alone or applying a stronger text encoder does not suffice to solve the task, and additional text tokens offer little benefit. Interestingly, we find that fine-grained visual reasoning is not exclusive to generative models trained by an autoregressive loss: When converted into CLIP-like encoders by contrastive finetuning, these MLLMs still outperform CLIP under the same cosine similarity-based evaluation protocol. Our study highlights the importance of VLM architectural choices and suggests directions for improving the performance of CLIP-like contrastive VLMs.
\end{abstract}
\section{Introduction}
\begin{figure}[t]
\begin{center}
\includegraphics[width=1.0\linewidth]{main_texts/intro.pdf}
\end{center}
\vspace{-10pt}
\caption{(a) Average two-way individual accuracy and pair accuracy of CLIP-ViT-L/14-336px and LLaVA-1.5-7B on various benchmarks~\citep{kamath2023s,hsieh2024sugarcrepe,thrush2022winoground,li2024naturalbench,yarom2024you,tong2024eyes}. (b) CLIP and Generative MLLM architectures (using LLaVA-1.5 as an example) for fine-grained visual reasoning tasks. We observe that Generative MLLMs perform better in extracting and utilizing query-relevant information from the same vision encoder. }
\label{fig1}
\vspace{-15pt}
\end{figure}
Despite the success and widespread adoption of Contrastive Language-Image Pretraining (CLIP)~\citep{radford2021learning}, recent studies have pointed out that state-of-the-art CLIP models still fall short in various visual reasoning tasks, including Winoground~\citep{thrush2022winoground}, SugarCREPE~\citep{hsieh2024sugarcrepe}, and What'sUp~\citep{kamath2023s}. These benchmarks require vision-language models (VLMs) to pair images and captions, which are carefully designed to test model capabilities of visio-linguistic compositional reasoning, spatial reasoning, or fine-grained detail understanding—areas beyond standard zero-shot classification on ImageNet. While CLIP excels at the latter, its performance in these visual reasoning tasks remains poor. 

One plausible explanation for these shortcomings is the potential information loss during the encoding process of the CLIP vision encoder~\citep{tong2024eyes}. For example, the encoder might behave like a bag-of-words model which only grasps the individual concepts in the image (``\texttt{mug}'' and ``\texttt{plate}'' in Figure~\ref{fig1}), but not the structural relationship (``\texttt{the mug is to the left of the plate}'')~\citep{yuksekgonul2023and}. 

In this work, we observe that the query-relevant visual information could still be preserved by CLIP vision encoder, but a better strategy is required to extract it: As shown in Figure~\ref{fig1}, LLaVA-1.5-7B~\citep{liu2024improved} with the \emph{same} pretrained vision encoder, surpasses CLIP-ViT-L/14-336px by a large margin on many challenging visual reasoning benchmarks. 
Particularly, on spatial reasoning benchmark What'sUp, while CLIP's pair accuracy is lower than random chance (25\%), LLaVA-1.5 achieves beyond 50\% on all four subsets (Table~\ref{tab:comparison-1}). More evidence of other Generative MLLMs on various benchmarks showing this phenomenon is presented in Section~\ref{sec:compare}.
These results indicate that these Generative MLLMs extract and utilize query-relevant information more effectively from the same CLIP vision encoder. Notably, the vision encoder remains unchanged throughout training, ensuring a fair comparison.

What is the driving force behind Generative MLLMs' extracting more visual information and achieving strong visual reasoning performance? How can it benefit and improve CLIP-like contrastive VLMs? In Section~\ref{ablation}, we investigate these questions by conducting controlled experiments on various factors as follows:
\begin{itemize}[noitemsep, left=0pt]
    \item \textbf{Training data.} In Section~\ref{ab1_data}, we observe little performance gain after directly finetuning CLIP on LLaVA-1.5's training data and hard negatives, indicating that training data is not the only contributor.
    \item \textbf{Token usage and position embedding.} In Section~\ref{ab2_token}, we observe that using patch tokens instead of the [CLS] token of CLIP (as proposed in PACL~\citep{mukhoti2023open}) brings improvement, and adding Rotary Position Embedding (RoPE)~\citep{su2024roformer} yields higher pair accuracy. However, using multiple text tokens from the CLIP text encoder as SPARC did~\citep{bica2024improving} does not help.
    \item \textbf{Language models.} In Section~\ref{ab3_lm_choice}, we replace the CLIP text encoder with a stronger, LLM-converted model~\citep{huang2024llm2clip}, but it does not suffice to realize effective extraction and outperform random chance. 
    \item \textbf{Architecture design for image-text alignment.} In Section~\ref{ab4_vlm2vec}, we find that text generation is not the only path to visual reasoning, as image-text matching through cosine similarity performed by contrastive VLMs can have strong performance on challenging benchmarks.
    \item \textbf{Training objective for image-text alignment.} In Section~\ref{ab4_vlm2vec}, we discover that finetuning with autoregressive loss is not necessary for deriving a VLM with fine-grained visual reasoning ability. 
    \item \textbf{Question as prompt.} In Section~\ref{ab4_vlm2vec}, we also investigate the role of the question as a prompt for Generative MLLMs and find that, when fully fused with the image, it reweights the image tokens, significantly aiding in the extraction of relevant information and the enhancement of image embeddings.
\end{itemize}
In Section~\ref{discussion_related_work}, we discuss the implications of our findings and their connection to prior work. Overall, we provide insights into VLM design and propose directions for improving contrastive VLMs. 
\section{Comparing CLIP and Generative MLLMs' visual reasoning performance}
\label{sec:compare}
We begin by introducing the task setup for the comparison. Using score-based evaluation, we notice a significant performance gap between CLIP and Generative MLLMs with the same vision encoder across several challenging visual reasoning benchmarks, highlighting the latter's stronger ability to extract and utilize visual information for reasoning.
\subsection{Task Setup}\label{background}
\begin{table*}[t]
\begin{center}
\vspace{-10pt}
\begin{tabular}{lcccccccc}
\toprule
& \multicolumn{4}{c}{What'sUp Subset A} & \multicolumn{4}{c}{What'sUp Subset B} \\
& \multicolumn{2}{c}{Left/Right} & \multicolumn{2}{c}{On/Under} & \multicolumn{2}{c}{Left/Right} & \multicolumn{2}{c}{Front/Behind}  \\
& Indiv. & Pairs & Indiv. & Pairs & Indiv. & Pairs & Indiv. & Pairs \\
\midrule
CLIP-ViT-L/14-336px & 49.0 & 1.9 & 61.7 & 23.3 & 54.9 & 10.8& 51.5& 7.8\\ 
\noalign{\vskip 0.3mm} \cdashline{1-9} \noalign{\vskip 0.3mm}
LLaVA-1.5-7B & 96.6& 93.2 & 76.2 & 52.4 & 98.5& 97.1 & \textbf{76.0}& \textbf{52.9}\\ 
Phi-3-V-3.8B & 97.6 & 95.1 & 78.6& 58.3 & \textbf{100} & \textbf{100} & 61.8& 26.5\\ 
LLaMA-3-V-8B & \textbf{98.1} & \textbf{96.1} & \textbf{81.1} & \textbf{64.1} & \textbf{100} & \textbf{100} & 73.0 & 47.1\\ 
\midrule 
Random chance & 50.0 & 25.0  & 50.0 & 25.0   & 50.0 & 25.0  & 50.0 & 25.0\\
\bottomrule
\end{tabular}
\vspace{-5pt}
\caption{The two-way individual accuracy and pair accuracy of CLIP-ViT-L/14-336px and Generative MLLMs in percentage points on four subsets of What'sUp. Generative MLLMs outperform CLIP by a large margin.}
\label{tab:comparison-1}
\end{center}
\vspace{-10pt}
\end{table*}
\begin{table*}[t]
\begin{center}
\begin{tabular}{lcccccccc}
\toprule
& Winoground & NaturalBench-R & MMVP & MMVP-VLM \\
\midrule
CLIP-ViT-L/14-336px & 27.8 & 47.8&14.0&20.7\\ 
\noalign{\vskip 0.3mm} \cdashline{1-5} \noalign{\vskip 0.3mm}
LLaVA-1.5-7B & 39.8 & 52.2  &36.0 & \textbf{49.6}\\ 
Phi-3-V-3.8B & 35.8& 50.5& 30.7& 31.9\\ 
LLaMA-3-V-8B & \textbf{46.3} & \textbf{64.7} & \textbf{50.0}&\textbf{49.6}\\ 
\midrule 
Random chance & 25.0  & 25.0   & 25.0  & 25.0\\
\bottomrule
\end{tabular}
\vspace{-5pt}
\caption{The pair accuracy of  CLIP-ViT-L/14-336px and Generative MLLMs in percentage points on several paired benchmarks. Generative MLLMs achieve substantially better performance than CLIP. }
\label{tab:comparison-2}
\end{center}
\vspace{-15pt}
\end{table*}
\setlength{\tabcolsep}{2pt}
\begin{table*}[t]
\begin{center}
\small
\vspace{-10pt}
\begin{tabular}{lccccccccccc}
\toprule
&SugarCREPE & SeeTrue &\textbf{What'sUp A} &\textbf{What'sUp B}&\multicolumn{2}{c}{\textbf{COCO-spatial}} & \multicolumn{2}{c}{\textbf{GQA-spatial}} \\
 &  & &    &  & One-obj. & Two-obj. & One-obj. & Two-obj. \\
\midrule 
CLIP-ViT-L/14-224px & 79.2 & 62.6   & 26.7 &25.7 & 49.1 & 50.2 & 46.0 & 48.1   \\
CLIP-ViT-L/14-336px & 80.0& 63.0  & 28.9 &27.2 & 48.9 & 51.1 & 46.6 & 49.1   \\
SigLIP-ViT-L/16-384px & 85.2 & 66.8   & 26.7& 28.7&50.3&48.6&47.8&48.7\\
EVA01-ViT-g-14 & 81.1&64.9 & 28.2 & 27.9&45.9&50.5&44.4&49.8\\
\noalign{\vskip 0.5mm} \cdashline{1-9} \noalign{\vskip 0.5mm}
LLaVA-1.5-7B &88.5&76.0& \textbf{69.9}&\textbf{65.4}&89.9&\textbf{88.9}&94.6&\textbf{95.2} \\
Phi-3-V-3.8B &82.8 &73.7 & 66.0 &52.7 & 89.5  & 79.8 & 93.0 & 87.3 \\
LLaMA-3-V-8B & \textbf{91.2} & \textbf{80.7} & 66.7& 58.6& \textbf{91.9}& 78.9  &\textbf{95.3 }&  91.4\\
\midrule 
Random chance & 50.0 & 50.0  & 25.0 & 25.0 & 50.0 & 50.0  & 50.0 & 50.0\\
\bottomrule
\end{tabular}
\vspace{-5pt}
\caption{Individual accuracy or AUROC of varied VLMs on visual reasoning benchmarks (spatial reasoning benchmarks in bold). The Generative MLLMs consistently outperform CLIP models (except on SugarCREPE, where SigLIP is better than Phi-3-V), with the largest performance gap observed in spatial reasoning.}
\label{tab:comparison-3}
\end{center}
\vspace{-15pt}
\end{table*}
This paper focuses on the image-text matching task in which VLMs are asked to choose from captions for a given image or vice versa. 

\noindent \textbf{Benchmarks.}
We use several challenging benchmarks, \textbf{Winoground}~\citep{thrush2022winoground}, \textbf{NaturalBench}~\citep{li2024naturalbench}, \textbf{SeeTrue}~\citep{yarom2024you}, \textbf{SugarCREPE}~\citep{hsieh2024sugarcrepe}, for assessing VLMs' compositionality. In Winoground, each test case has two image+text pairs with the same words in different order. For NaturalBench, we use the retrieval version (denoted as NaturalBench-R) in the same format as Winoground provided by \citet{lin2024evaluating}. SeeTrue consists of individual image-text pairs, while SugarCREPE has one image and two captions per test case. We use \textbf{MMVP(-VLM)}~\citep{tong2024eyes} to test VLMs' ability to capture visual details like object existence, orientation, and counting. Since MMVP is not in paired image-text format, we manually convert it without altering content. We adopt \textbf{What'sUp A\&B} with \textbf{COCO-spatial} and \textbf{GQA-spatial}~\citep{kamath2023s} to evaluate VLMs' spatial reasoning. For What'sUp, each test case includes four captions (e.g., ``\texttt{A dog left of/right of/on/under a table}'') and corresponding images with minimal variation except for spatial relationships. We split each test case into two pairs—e.g., one pair contrasts ``\texttt{left of}'' versus ``\texttt{right of}'' with their ground truth images, and the other covers the remaining captions. This yields four benchmark subsets for A and B. COCO-spatial and GQA-spatial have one image and two captions per test case. More details are in Appendix~\ref{appendix_benchmark}.

\noindent \textbf{Models.} Our main comparison is between \textbf{CLIP-ViT-L/14-336px}~\citep{radford2021learning} and Generative MLLMs that use its pretrained vision encoder and keep the weights frozen during training: \textbf{LLaVA-1.5-7B}~\citep{liu2024improved}, along with \textbf{Phi-3-V-3.8B} and \textbf{LLaMA-3-V-8B}~\citep{hanoona2024LLaVA++}. In these MLLMs, the patch tokens from the CLIP vision encoder first pass through a two-layer MLP connector and are then used as input tokens for a generative language model which yields the token probability determining the model response. We also include results of \textbf{CLIP-ViT-L/14-224px}, \textbf{SigLIP-ViT-L/16-384px}~\citep{zhai2023sigmoid}, and \textbf{EVA01-ViT-g-14}~\citep{sun2023eva} for reference since they are of interest and widely used~\citep{tong2024cambrian}.

\noindent \textbf{Evaluation protocol.} For CLIP-like contrastive VLMs, the matching score is the cosine similarity between its image embeddings and text embeddings. In prior works, Generative MLLMs are commonly evaluated by GPT-4~\citep{achiam2023gpt} or human evaluators on generated responses. However, human evaluators are expensive for thousands of model responses, and GPT-4 as the judge can be incorrect and affected by user prompts. To ensure a fair comparison, we choose to use a score-based evaluation method and adopt the VQAScore~\citep{lin2024evaluating}, defined as  
\begin{align*}
    P (\text{``Yes''}|\text{\texttt{image}, ``Does this figure show `{\texttt{text}}'? }\\
    \text{Please answer yes or no.''})
\end{align*}
The question template remains the same across different benchmarks. We present the comparison between VQAScore and response-based evaluation in Appendix~\ref{appendix_response}.

\noindent \textbf{Evaluation metrics.} For SeeTrue, we report an average AUROC of three subsets. For other benchmarks, we use pair accuracy and individual accuracy when applicable. \textbf{Pair accuracy}~\citep{tong2024eyes, kamath2023s} requires correct matching for both images, and it only applies to benchmarks with two images in a test case. \textbf{Individual accuracy} refers to the accuracy of individual images. For MMVP(-VLM), we follow the original paper and use pair accuracy to represent the correct matching for both \emph{captions} and individual accuracy for individual \emph{captions} instead.
\subsection{Results}\label{blindness}
We present the comparison in Table~\ref{tab:comparison-1},~\ref{tab:comparison-2}, and~\ref{tab:comparison-3}. On these challenging benchmarks, Generative MLLMs outperform CLIP-ViT-L/14-336px with the same vision encoder, showing that (1) CLIP vision encoder has much query-relevant visual information not utilized by CLIP, and (2) Generative MLLMs can extract and align this information from the encoder more effectively. The performance gap is the most significant on \textbf{spatial reasoning}, where the CLIP models behave close to random chance for individual accuracy and lower than random chance for pair accuracy, but Generative MLLMs achieve high accuracies. We further find that the Generative MLLMs can even outperform XVLM~\citep{zeng2021multi} specialized in spatial reasoning (See Appendix~\ref{appendix_xvlm}).
\section{Investigation of the Performance Gap}\label{ablation}
\begin{figure}[t]
\begin{center}
\includegraphics[width=1.0\linewidth]{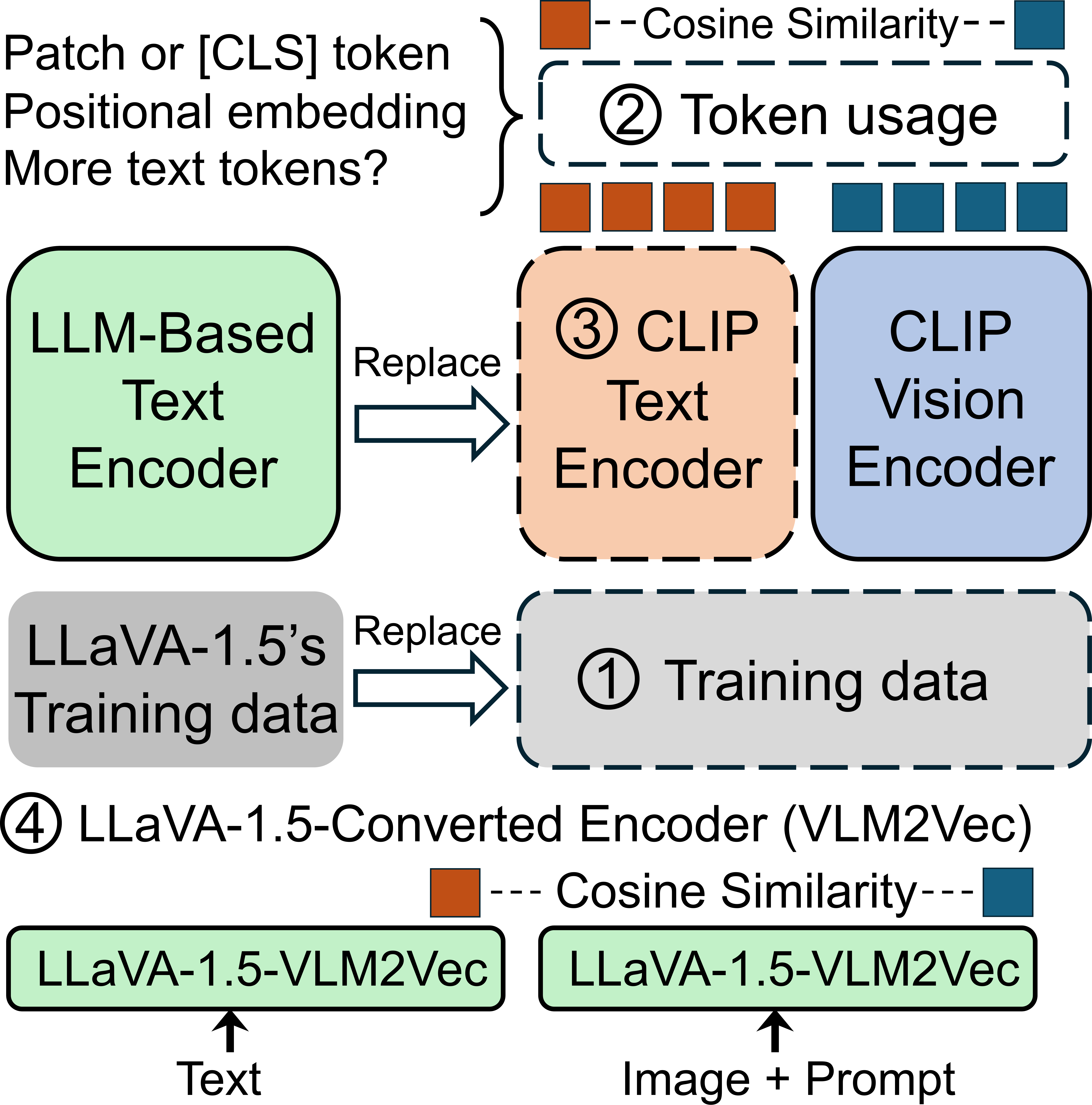}
\end{center}
\vspace{-10pt}
\caption{An illustration for CLIP-like contrastive VLMs and the controlled experiments in Section~\ref{ablation}. We first investigate the effect of training data by replacing them with LLaVA-1.5's training data ($\circled{1}$). Then, we try different token usage for CLIP vision encoder and text encoder ($\circled{2}$) and discuss the influence of using stronger text encoders converted from LLMs ($\circled{3}$). Finally, we convert LLaVA-1.5 to contrastive VLMs ($\circled{4}$) to study the effect of the alignment architecture and training objective.}
\label{roadmap}
\vspace{-20pt}
\end{figure}

The gap observed in Section~\ref{blindness} could be the result of various factors, ranging from model training to architecture. In this section, we try to dissect and examine which factors contribute to Generative MLLMs' success and cause CLIP's failure by controlled experiments. We focus on the performance gap on \textbf{What'sUp}, of which the test cases are tightly controlled and balanced. A road map of the experiments is illustrated in Figure~\ref{roadmap}.
\subsection{Training Data}\label{ab1_data}
\setlength{\tabcolsep}{2pt}
\begin{table}[t]
\begin{center}
\small
\begin{tabular}{lccccccccc}
\toprule
& \multicolumn{2}{c}{What'sUp Subset A}&\multicolumn{2}{c}{What'sUp Subset B} \\
 & Indiv. & Pairs & Indiv. & Pairs \\
\midrule 
CLIP & 49.0&1.9&54.9&10.8\\
\small{+ finetuning (ft)}& 50.5&1.9&53.9&5.9\\
\small{+ ft + hard neg. } & 50.5&1.0&50.5&1.0 \\
SigLIP &50.0&1.9&51.5&5.9\\
\small{+ finetuning (ft)}& 49.0&1.0&51.0&3.9\\
\small{+ ft + hard neg. } &50.0&0.0&50.0&0.0\\
EVA-CLIP & 49.0&1.0& 50.1&4.9\\
\small{+ finetuning (ft)} & 50.0& 4.9& 48.5& 2.0\\
\small{+ ft + hard neg. } & 50.0& 1.9& 48.0&2.0\\
\midrule 
Random chance & 50.0 & 25.0  & 50.0 & 25.0  \\
\bottomrule
\end{tabular}
\vspace{-5pt}
\caption{The two-way individual accuracy and pair accuracy results of CLIP-ViT-L/14-336px, SigLIP-ViT-L/16-384px, and EVA01-ViT-g-14 focusing on the \textbf{Left/Right} subsets of What'sUp after finetuning on LLaVA-1.5's training data with or without hard negative captions. After direct finetuning, the accuracies are still quite low.}
\label{tab:ablation_data}
\end{center}
\vspace{-20pt}
\end{table}
First, we hypothesize that Generative MLLMs' visual information extraction ability benefits from training data. To check the effect of data, we use LLaVA-1.5's training data to finetune CLIP, SigLIP, and EVA-CLIP. We convert the datasets to the image-caption format (Details are deferred to Appendix~\ref{appendix_data}). By default, we freeze the vision encoder during finetuning for strict ablation. Considering that contrastive learning relies on negative samples beyond data quality~\citep{robinson2020contrastive, kalantidis2020hard}, we also construct hard negative captions by switching the related phrases to their opposite (e.g., replacing ``\texttt{on the left}'' with ``\texttt{on the right}''). In this setting, the training objective follows NegCLIP~\citep{yuksekgonul2023and}. 

Results are shown in Table~\ref{tab:ablation_data}. Finetuning on LLaVA-1.5's training data does not help these models, even with hard negatives. Still, their accuracy is around random chance. We also try to unlock the SigLIP vision encoder during finetuning, which does not increase the performance either (See results in Appendix~\ref{appendix_unlock}). We experiment with XVLM~\citep{zeng2021multi} and observe similar results in Appendix~\ref{appendix_xvlm}. This finding aligns with the previous failure on finetuning them on a much larger, preposition-focused subset of LAION~\citep{kamath2023s}, indicating that \textbf{data alone does not lead to stronger extraction ability.}
\subsection{Token Usage}\label{ab2_token}
\setlength{\tabcolsep}{2pt}
\begin{table*}[t]
\begin{center}
\vspace{-10pt}
\begin{tabular}{lcccccccc}
\toprule
& \multicolumn{4}{c}{What'sUp Subset A} & \multicolumn{4}{c}{What'sUp Subset B} \\
& \multicolumn{2}{c}{Left/Right} & \multicolumn{2}{c}{On/Under} & \multicolumn{2}{c}{Left/Right} & \multicolumn{2}{c}{Front/Behind}  \\
& Indiv. & Pairs & Indiv. & Pairs & Indiv. & Pairs & Indiv. & Pairs \\
\midrule
LLaVA-1.5-7B-LoRA & \textbf{84.5}& \textbf{68.9}& \textbf{76.2}& \textbf{52.4}& \textbf{89.2}& \textbf{78.4}&\textbf{86.3} &\textbf{72.5}\\ 
\texttt{[CLS]}-LLaVA-1.5-7B-LoRA & 44.2& 8.7& 54.4& 8.7& 49.0& 4.9&53.9 &12.7\\
\midrule 
Random chance & 50.0 & 25.0  & 50.0 & 25.0 & 50.0 & 25.0  & 50.0 & 25.0  \\
\bottomrule
\end{tabular}
\vspace{-5pt}
\caption{The results of \texttt{[CLS]}-LLaVA-1.5-7B-LoRA and reproduced LLaVA-1.5-7B-LoRA on all subsets of What'sUp, where \texttt{[CLS]}-LLaVA-1.5-7B-LoRA struggles with spatial reasoning. }
\label{tab:ablation_cls_llava}
\end{center}
\vspace{-10pt}
\end{table*}
\setlength{\tabcolsep}{2pt}
\begin{table*}[t]
\begin{center}
\begin{tabular}{lccccccccc}
\toprule
& \multicolumn{4}{c}{What'sUp Subset A} & \multicolumn{4}{c}{What'sUp Subset B} \\
& \multicolumn{2}{c}{Left/Right} & \multicolumn{2}{c}{On/Under} & \multicolumn{2}{c}{Left/Right} & \multicolumn{2}{c}{Front/Behind}  \\
& Indiv. & Pairs & Indiv. & Pairs & Indiv. & Pairs & Indiv. & Pairs \\
\midrule 
CLIP-ViT-L/14-336px & 49.0 & 1.9 & \textbf{61.7} & \textbf{23.3} & \textbf{54.9} & 10.8& 51.5& 7.8\\ 
+ Patch Tokens (PT)& 47.6& 9.7& 52.9& 10.7& 52.9& 9.8& 51.5& 6.9\\
+ PT + RoPE & \textbf{54.9}&\textbf{22.3} & 46.1& 13.6& 52.0& \textbf{20.6}&45.6 &12.7 \\
+ PT + RoPE + Multiple Text Tokens & 48.1& 0.0& 50.0&2.9 & 50.0&6.9 &48.0 & 7.8\\
+ PT + RoPE + Stronger Text Encoder & 50.5& 10.7&48.5& 6.8& 50.0& 15.7& \textbf{53.9}& \textbf{21.6}\\
{\color{gray}{LLM2CLIP~\citep{huang2024llm2clip}}} &{\color{gray}{49.5}} & {\color{gray}{1.0}}& {\color{gray}{58.7}}& {\color{gray}{17.4}}& {\color{gray}{49.0}}& {\color{gray}{1.0}}& {\color{gray}{55.4}}& {\color{gray}{14.7}}\\
\midrule 
Random chance & 50.0 & 25.0  & 50.0 & 25.0 & 50.0 & 25.0  & 50.0 & 25.0  \\
\bottomrule
\end{tabular}
\vspace{-5pt}
\caption{The results of different token usage and leveraging a stronger text encoder for CLIP-ViT-L/14-336px on the What'sUp benchmark after finetuning on LLaVA-1.5's training data. CLIP with Patch tokens + RoPE has the highest average pair accuracy.}
\label{tab:ablation_pacl}
\end{center}
\vspace{-15pt}
\end{table*}
\textbf{Patch tokens.} The output of the CLIP vision encoder consists of two parts: The \textbf{\texttt{[CLS]} token}, functioning as the global feature of the image, and the \textbf{patch tokens}, containing local information of image patches. We notice that these Generative MLLMs employ all 576 patch tokens from the CLIP-ViT-L/14-336px vision encoder, in contrast to CLIP using only the projected \texttt{[CLS]} token. 

We first perform an ablation study on LLaVA-1.5: We change the input of its language model to use only the \texttt{[CLS]} token, train this ``\texttt{[CLS]}-LLaVA-1.5'' model from scratch (pretraining + finetuning) using LoRA~\citep{hu2021lora}, and observe that its spatial reasoning performance is significantly worse than our reproduced LLaVA-1.5-LoRA in Table~\ref{tab:ablation_cls_llava}. This proves the importance of patch tokens to fine-grained visual reasoning: \textbf{Detailed information of images resides in these patch tokens.} 

Inspired by this finding, we try incorporating patch tokens in standard CLIP models. We adopt the PACL method~\citep{mukhoti2023open} as it proposes to train a vision embedder $e_v$ for patch tokens and a text embedder $e_t$ on top of the frozen CLIP model (consisting of vision encoder $f_v$ and text encoder $f_t$). For input image $\mathbf{x}$ and text $\mathbf{y}$, we calculate the image feature $\mathbf{v}(\mathbf{x})$ by
\begin{align*}
    s(\mathbf{x}, \mathbf{y}) &= e_v(f_v(\mathbf{x}))\cdot e_t(f_t(\mathbf{y})) \\
    \mathbf{v}(\mathbf{x}) &= e_v(f_v(\mathbf{x}))^\top \cdot\text{sigmoid}(10\cdot s(\mathbf{x}, \mathbf{y}))
\end{align*}
In other words, $s(\mathbf{x}, \mathbf{y})$ determines the weight for each projected patch token based on the text, and $\mathbf{v}(\mathbf{x})$ is a weighted sum of all projected patch tokens.
Then we use $\mathbf{v}(\mathbf{x})$ and $e_t(f_t(\mathbf{y}))$ as the image and text features for CLIP training with the original contrastive objective.
During the evaluation, we use the average of projected patch tokens $e_v(f_v(\mathbf{x}))$ as the image feature and $e_t(f_t(\mathbf{y}))$ as the text feature. The results of training on LLaVA-1.5's data are shown in the second row of Table~\ref{tab:ablation_pacl}. It brings higher pair accuracy to the Left/Right subset in Subset A. 

\noindent \textbf{Position embeddings.} Considering that the average or weighted sum does not maintain the order/positional information of patch tokens, we add Rotary Position Embeddings (RoPE)~\citep{su2024roformer} to $f_v(\mathbf{x})$ before passing it to the vision embedder $e_v$, since RoPE is applied to visual tokens in the language model of Generative MLLMs we study. In the third row of Table~\ref{tab:ablation_pacl}, we find that this combination yields significantly higher pair accuracy on three subsets, showing that \textbf{part of the information comes from the order of patch tokens.} Nonetheless, the individual accuracy is not improved by much.

\noindent \textbf{Multiple text tokens.} Does using multiple text tokens of CLIP text encoder as well offer further performance gain? In Generative MLLMs, it is natural to use multiple visual tokens and text tokens, as they are concatenated as the input of an autoregressive language model. However, it is non-trivial to do so in contrastive VLMs. Therefore, we leverage the SPARC method~\citep{bica2024improving} to implement the interaction between multiple visual tokens and text tokens for CLIP: We first obtain a weighted sum of patch tokens for each text token (named grouped visual tokens) and then perform local contrastive learning between grouped visual tokens and text tokens within each sample. The training objective is the sum of this local contrastive loss and the standard contrastive loss. For evaluation, we use the average of grouped visual tokens as the image feature and the average of text tokens as the text feature. Details are deferred to Appendix~\ref{appendix_sparc}. Despite the complexity, this method does not help our task (See the fourth row of Table~\ref{tab:ablation_pacl}). This failure might result from the hardness of training and the insufficiency of token interaction.

\subsection{Language Model}\label{ab3_lm_choice}
Previous research suggests that the CLIP text encoder fails to capture changed word orders, negation, and spatial or numerical details~\citep{tong2024mass, kamath2023text, yuksekgonul2023and}, while Generative MLLMs employ powerful pretrained LLMs, which is supposed to be stronger than the CLIP text encoder at reasoning. 

Are pretrained LLMs the missing piece to effectively extracting visual information? We perform further experiments on finetuning CLIP with patch tokens and RoPE on LLaVA-1.5 training data but replacing the original CLIP text encoder with a stronger one provided by LLM2CLIP~\citep{huang2024llm2clip}. This text encoder is converted from Llama-3-8B-Instruct~\citep{dubey2024llama} by contrastive finetuning and is shown to bring performance boost to state-of-the-art CLIP models on benchmarks such as MS COCO~\citep{lin2014microsoft}. We keep this text encoder and the CLIP vision encoder frozen during our finetuning. The results are shown in the fifth row of Table~\ref{tab:ablation_pacl}, where we also attach the results of the original LLM2CLIP checkpoint of CLIP-ViT-L/14-336px for reference\footnote{The original LLM2CLIP is not a fair comparison as its implementation unfreezes the CLIP vision encoder during finetuning.}. We find that \textbf{a stronger text encoder does not suffice to effectively extract more information towards solving the task.}
\setlength{\tabcolsep}{2pt}
\begin{table*}[t]
\begin{center}
\vspace{-10pt}
\begin{tabular}{lcccccccc}
\toprule
& \multicolumn{4}{c}{What'sUp Subset A} & \multicolumn{4}{c}{What'sUp Subset B} \\
& \multicolumn{2}{c}{Left/Right} & \multicolumn{2}{c}{On/Under} & \multicolumn{2}{c}{Left/Right} & \multicolumn{2}{c}{Front/Behind}  \\
& Indiv. & Pairs & Indiv. & Pairs & Indiv. & Pairs & Indiv. & Pairs \\
\midrule
CLIP-ViT-L/14-336px & 49.0 & 1.9 & 61.7 & 23.3 & 54.9 & 10.8& 51.5& 7.8\\ 
\noalign{\vskip 0.3mm} \cdashline{1-9} \noalign{\vskip 0.3mm}
LLaVA-1.5-7B-VLM2Vec-LoRA & \textbf{97.1} & \textbf{95.1} & \textbf{68.0} & \textbf{35.9} & \textbf{100}&  \textbf{100}&  \textbf{60.8}& \textbf{22.5}\\
w/o Question in Prompt & 49.5 & 0.0 & 50.5 & 1.9 & 46.6 & 2.0 & 50.5 & 1.0\\
\midrule 
Random chance & 50.0 & 25.0  & 50.0 & 25.0 & 50.0 & 25.0  & 50.0 & 25.0  \\
\bottomrule
\end{tabular}
\vspace{-5pt}
\caption{The two-way individual accuracy and pair accuracy of CLIP-ViT-L/14-336px and LLaVA-1.5-converted models in percentage points on four subsets of What'sUp. LLaVA-1.5-7B-VLM2Vec-LoRA outperforms CLIP on all subsets. When there is no question in the prompt, its performance degenerates to the standard CLIP.}
\label{tab:vlm2vec}
\end{center}
\vspace{-15pt}
\end{table*}

\subsection{Alignment Architecture, Training Objective, and Prompt}\label{ab4_vlm2vec}
A major difference between CLIP-like contrastive VLMs and LLaVA-like Generative MLLMs is how they align images and texts. However, it is hard to examine every factor involved separately: The alignment architecture of CLIP—cosine similarity between image embeddings and text embeddings—is bound to its training objective (contrastive loss) and contrastive VLM structure (dual encoders). On the other hand, it is plausible to hypothesize that contrastive VLMs cannot perform fine-grained visual reasoning since cosine similarity might be overly coarse-grained both for training and evaluation, compared with text generation and autoregressive loss used by Generative MLLMs.

We bypass this obstacle in comparison by converting a Generative MLLM to a CLIP-like contrastive VLM. On LLaVA-1.5-7B, we use the converting method proposed by VLM2Vec~\citep{jiang2024vlm2vec}: Specifically, we take the last layer vector representation of the last output token of LLaVA-1.5-7B as the output embedding. In this way, we get the encoder for image+question(prompt) and pure text simultaneously since LLaVA-1.5 allows using one or zero images in the input. Following the original paper, we use the prompt templates: ``\texttt{Represent the given image with the following question: \{Question\}}'' while encoding the image if there is a question in the sample; ``\texttt{Find the text that can answer the given query: \{Question\}}'' when there is no image; and no additional prompt for encoding im-
age+question of LCS-558K and the captions. Then, we finetune this encoder using contrastive loss and LoRA~\citep{hu2021lora} on LLaVA-1.5's training data with the CLIP vision encoder frozen. Surprisingly, they exhibit strong performance without using a large batch size (256) in Table~\ref{tab:vlm2vec} (LLaVA-1.5-7B-VLM2Vec-LoRA). The question used for evaluation is listed in Appendix~\ref{appendix_eval_vlm2vec}. \textbf{This proves that text generation+autoregressive loss is not the only solution to fine-grained visual reasoning. }

What could be the key factor of the success of this contrastive LLaVA-1.5 compared with CLIP models, including the standard ones and ours with patch tokens plus RoPE in Section~\ref{ab2_token}? We verify that the additional question added in the prompt when obtaining the image embeddings plays an important role here. When we change the prompt template to ``\texttt{Represent the given image.}'' without any question, the model performance degenerates to the standard CLIP performance as shown in the third row of Table~\ref{tab:vlm2vec}. Therefore, we conclude that \textbf{the question greatly helps the extraction and utilization of visual information from the vision encoder.} The question helps to reweight the patch tokens according to the context. Without the question, the image embeddings remain the same regarding different tasks (e.g., coarse-grained classification like ``\texttt{dog/cat}'', versus fine-grained visual reasoning like ``\texttt{dog to the left/right of the table}''), which could be suboptimal and cause difficulty in alignment.
\section{Discussion and Connection to Prior Work}\label{discussion_related_work}
In this section, we first discuss how our findings connect to the observations and conclusions in existing literature. Then, we list two directions for improving VLM's visual reasoning ability based on our results. 
\subsection{Connection to Prior Work}
Recent results of VLMs on various benchmarks for testing fine-grained visual reasoning ability (e.g., compositionality, spatial reasoning, counting) reveal that they fail to solve simple tasks unexpectedly and often ignore visual patterns in the image~\citep{thrush2022winoground, yuksekgonul2023and}. Researchers are actively exploring the root causes of such failures. \citet{lin2024evaluating} notices the advantage of Generative MLLMs over CLIP in image-text matching tasks. We observe the significant discrepancy in performance when controlling the vision encoder and thus focus on how Generative MLLMs could outperform CLIP-like contrastive VLMs with the same vision encoder.

\noindent \textbf{Vision encoder and token usage.} \citet{tong2024eyes} observes that the CLIP vision encoder could encode visually distinct images into highly similar embeddings, omitting essential information and thus resulting in low accuracy on tasks regarding the visual semantic difference. Hence, they suggest using features from multiple vision encoders, which is adopted by later works~\citep{kar2024brave, tong2024cambrian, xu2024libra}. However, we observe that this part of information could be captured by the CLIP vision encoder but is not extracted or aligned properly. Similar to our observation, \citet{koishigarina2025clip} argues that CLIP is not bag-of-words uni-modally, and the real issue of CLIP's compositionality lies in poor cross-modal alignment. Besides, while \citet{tong2024eyes} only calculates the similarity between \texttt{[CLS]} tokens used by CLIP as evidence, we argue that detailed information is preserved in patch tokens and their positions. 

\noindent \textbf{Text encoder.} \citet{kamath2023text} and \citet{tong2024mass} point out that the CLIP text encoder might discard relevant information during encoding so that the model could not discriminate images that differ in key aspects. Following previous efforts in converting LLM to an encoder~\citep{behnamghader2024llm2vec}, recent works explore using LLM-converted encoders as the text encoder for CLIP: LLM2CLIP~\citep{huang2024llm2clip} finds that this practice boosts performance on several retrieval tasks on top of state-of-the-art CLIP models, but we observe its unsatisfying performance on What'sUp; \citet{stone2024learning} achieves high accuracy on challenging benchmarks for compositionality after large-scale pretraining, although they reported struggles on Left/Right spatial relations. We discover that a stronger text encoder is not enough for solving the fine-grained visual reasoning task.

\noindent \textbf{Training data and objective.} Data-centric methods for improving CLIP-like models include selecting or synthesizing higher-quality image-text pairs~\citep{gadre2024datacomp, nguyen2024improving, zheng2024dreamlip}, involving more negative samples by manual design~\citep{yuksekgonul2023and, paiss2023teaching} or larger batch size~\citep{stone2024learning}. But \citet{kamath2023s} observes that CLIP cannot learn spatial relations even after training on a large amount of relevant data, suggesting that we might need inductive bias or denser supervision like XVLM~\citep{zeng2021multi}. Others try applying autoregressive loss, such as Cap/CapPa~\citep{tschannen2023image}, or combining it with contrastive loss, like CoCa~\citep{yu2022coca}. Inspired by VLM2Vec~\citep{jiang2024vlm2vec}, we train LLaVA-1.5-VLM2Vec and verify that task-specific inductive bias, additional supervision, manually designed hard negatives, or finetuning with autoregressive loss is not necessary for contrastive VLMs to learn spatial relations. 

\noindent \textbf{Alignment architecture of contrastive VLMs.} Cross-modal alignment can be implemented by cross-modal matching through cosine similarity~\citep{radford2021learning}, matching by directly outputting a score~\citep{li2023blip2}, and generation (outputting a response)~\citep{liu2024improved, awadalla2023openflamingo}. The cross-modal contrasting is efficient, yet unable to perform complex reasoning like the generative models where Chain-of-Thought is applicable~\citep{wei2022chain}. Nevertheless, our LLaVA-1.5-VLM2Vec experiments show that advanced techniques can ignite the potential of contrastive VLMs in visual information extraction and improve their visual reasoning performance. 

\subsection{Discussion on Improving VLMs' Visual Reasoning Ability}

\textbf{Promptable image embeddings boost performance on fine-grained tasks.} CLOC~\citep{chen2024contrastive} formulates the idea of promptable embedding for regional understanding of images. They pass image embeddings and spatial hints to a prompter for obtaining region representations of images and perform localized contrastive training. In this way, when a grounding task only requires information from part of the image, the representation will not be distracted by other parts and thus lead to higher accuracy. VLM2Vec~\citep{jiang2024vlm2vec} extends the spatial hints to general prompts and proposes a method for converting Generative MLLMs to encoders. Our ablation study of questions in prompts for LLaVA-1.5-VLM2Vec demonstrates the effectiveness of this technique.

\noindent \textbf{Effectively utilizing vision encoders offers benefits without pretraining new vision models.} Our results suggest that there is still room to enhance VLMs with a fixed, pretrained vision encoder by advanced extraction methods. We explore whether this also holds for Generative MLLMs in the Appendix~\ref{appendix_m3id}:
We try an alternative decoding algorithm on LLaVA-1.5-7B for attending more to the visual information, named Multi-Modal Mutual-Information Decoding~\citep{favero2024multi}, which leads to performance gain (+6\%), on par with using interleaved visual tokens from multiple vision encoders (I-MoF~\citep{tong2024eyes}). This result indicates that LLaVA-1.5 still misses some key information for query answering and has room for further improvement apart from using a better vision encoder. 
\section{Conclusion}
Our study first reveals that Generative MLLMs perceive fine-grained visual information more effectively using the same vision encoder than CLIP for visual reasoning tasks. Through controlled experiments, we find that patch tokens, position embeddings, and prompt-based image embeddings are key differences causing the gap; however, training data, multiple text tokens, and better text encoders are insufficient to bridge the gap. Additionally, text generation and finetuning with autoregressive loss are not mandatory for strong visual reasoning. These findings not only offer insights into VLM design but also provide practical guidelines for enhancing contrastive VLMs on visual reasoning.
\section{Limitations}
First, for controlled experiments on data in Section~\ref{ab1_data}, we do not train models from scratch or use larger batch sizes due to the limited computing resources, so the conclusion regarding data might be restricted.

Second, the number of visual reasoning benchmarks we study is restricted. Therefore, we hope that more comprehensive, unbiased, and visual-centric reasoning benchmarks for VLMs can be available in the future.

Third, we only study the comparison between CLIP-ViT-L/14-336px and the Generative MLLMs that use its vision encoder and explore the reasons behind their discrepancy. Our conclusion is thus restricted to them. We do not claim that all Generative MLLMs are better than contrastive VLMs in all cases. Nevertheless, it is interesting to compare other pairs of contrastive VLMs and Generative MLLMs, and we leave this for future work.

\section*{Acknowledgments}
We thank Amita Kamath for the discussion about the What'sUp benchmark. We appreciate the comments and advice from Rui Xin and Scott Geng on our drafts.
PWK is supported by the Singapore National Research Foundation and the National AI Group in the Singapore Ministry of Digital Development and Innovation under the AI Visiting Professorship Programme (award number AIVP-2024-001) and by the AI2050 program at Schmidt Sciences.
SSD acknowledges the support of  NSF DMS 2134106, NSF CCF 2212261, NSF IIS 2143493, NSF IIS 2229881, Sloan Fellowship, and the AI2050 program at Schmidt Sciences.

\bibliography{custom}

\appendix
\twocolumn[\clearpage]

\section{Benchmarks and Additional Evaluations}
\begin{figure}[t]
\begin{center}
\includegraphics[width=0.98\linewidth]{main_texts/Figure3.pdf}
\end{center}
\vspace{-10pt}
\caption{Example test case and evaluation method for CLIP-like models on What'sUp benchmark. In our two-way evaluation on benchmarks with paired images, a test case consists of two similar images and two captions. The model chooses one caption for each image, and it gets one point in pair accuracy only if choosing correctly for both images. The choices of CLIP-like models are determined by $S_C$, cosine similarity between image and text embeddings.}
\label{fig:example}
\vspace{-10pt}
\end{figure}
\subsection{Benchmark Information}\label{appendix_benchmark}
\paragraph{What'sUp.} The What'sUp benchmark~\citep{kamath2023s} contains 820 images of pairs of household objects captured by the authors, 408 in Subset A and 412 in Subset B. For every object pair, all prepositions are present in the benchmark, and thus the images and captions are balanced, avoiding the bias in real-world images (e.g., a cup is usually on the table, not under the table). We corrected the mislabeled images in the GitHub Issues and reevaluated the pretrained VLMs. For CLIP and XVLM's evaluation, we refer to the official code provided by the What'sUp benchmark's authors in \url{https://github.com/amitakamath/whatsup_vlms}. The evaluation of SigLIP and EVA-CLIP directly follows the evaluation of CLIP in the official code. We offer an example in Figure~\ref{fig:example} to demonstrate how pair accuracy and individual accuracy are computed on benchmarks with paired images like What'sUp.
\paragraph{COCO-spatial and GQA-spatial.} \citet{kamath2023s} also selects validation sample from COCO~\citep{lin2014microsoft} and GQA~\citep{hudson2019gqa} targeting spatial relations (to the left of vs to the right of, above vs below). Each test case contains one image, one positive caption, and one negative caption. COCO-spatial has 2687 test cases, and GQA has 1451 test cases in total.
\paragraph{Winoground, NaturalBench, and SeeTrue.} Winoground~\citep{thrush2022winoground} is a challenging benchmark consisting of 400 pairs of image-text pairs. It focuses on VLM's compositionality, with two images and two similar captions in one test case. A example of the captions is ``\texttt{some plants surrounding a lightbulb}'' vs ``\texttt{a lightbulb surrounding some plants}.'' High pair accuracy requires VLM to match these images with their captions correctly at the same time. NaturalBench~\citep{li2024naturalbench} is a benchmark for testing Generative MLLMs on compositionality with unbiased Yes/No answers. In one test case, there are two images with two questions, and each question has "Yes" as the answer for one image and "No" for the other image. We use the retrieval version of NaturalBench provided by~\citep{lin2024evaluating}. SeeTrue~\citep{yarom2024you} is an alignment bench that has 6930 human labels for whether a given image is paired with the text or not. We report the AUROC (Area Under the Receiver Operating Characteristic curve) instead of accuracy on SeeTrue. We use VQAScore's official code for evaluation on these benchmarks in \url{https://github.com/linzhiqiu/t2v_metrics}.
\paragraph{SugarCREPE.} SugarCREPE~\citep{hsieh2024sugarcrepe} is designed for evaluating VLM's compositionality with grammatical, sensical, and fluent hard negatives. Each test case contains one image, one positive caption, and one negative caption. There are 7512 test cases in total.
\paragraph{MMVP(-VLM).} The MMVP benchmark contains 150 pairs of similar images, and the MMVP-VLM benchmark has 135 pairs of similar images, divided into nine categories. There is an overlap between the image pairs in these two benchmarks. We corrected the mislabeled images in the GitHub Issues and reevaluated the pretrained VLMs. Since MMVP is incompatible with CLIP, we convert its questions manually. We attach the converted version to the supplementary material for reference.
\subsection{Model Weight Information}\label{appendix_model}
We use public pretrained weights of LLaVA-1.5-7B (\url{https://huggingface.co/llava-hf/llava-1.5-7b-hf}) under the Meta LLaMA License Agreement and the weights of Phi-3-V-3.8B in \url{https://huggingface.co/MBZUAI/LLaVA-Phi-3-mini-4k-instruct} and LLaMA-3-V-8B in \url{https://huggingface.co/MBZUAI/LLaVA-Meta-Llama-3-8B-Instruct} provided by~\citep{ranasinghe2024learning} under MIT License since they are trained with vision encoder frozen. For contrastive VLMs, we use OpenAI's pretrained CLIP-ViT-L/14-224px and CLIP-ViT-L/14-336px model under MIT License, SigLIP-ViT-L/16-384px pretrained on the WebLI dataset~\citep{chen2022pali} and EVA01-ViT-g-14 pretrained on the LAION400M-s11b-b41k dataset~\citep{schuhmann2021laion} under Apache 2.0 License provided in the OpenCLIP repository. In Table~\ref{tab:ablation_cls_llava}, LLaVA-1.5-7B-LoRA is reproduced. In Table~\ref{tab:ablation_pacl}, the checkpoint for LLM2CLIP is from \url{https://huggingface.co/microsoft/LLM2CLIP-Llama-3-8B-Instruct-CC-Finetuned} under Apache 2.0 License. We also use the text encoder and the adapter of this checkpoint in our experiments of using a stronger text encoder. In Table~\ref{tab:vlm2vec}, the LLaVA-1.5-7B-VLM2Vec-LoRA is trained by ourselves with the vision encoder frozen using the VLM2Vec method~\citep{jiang2024vlm2vec}.
\subsection{Comparison between VQAScore and Response-Based Evaluation}\label{appendix_response}
We compare the score-based evaluation, VQAScore~\citep{lin2024evaluating}, and the standard response-based evaluation for Generative MLLMs on What'sUp.
Response-based evaluation requires a question accompanied by a given image as the input, and the questions used for LLaVA-1.5's evaluation are listed in Table~\ref{tab:prompts}. Then, the question is concatenated with the fixed prompt template (``\texttt{USER: <image>\textbackslash n\{question\} ASSISTANT:}''). Considering the position bias in LLMs~\citep{wang2024eliminating}, we exchange the position of two prepositions in the question with 50\% probability on COCO-spatial and GQA-spatial benchmarks for fair results. On the What'sUp benchmark, the orders are always the same for two images. Then, we use greedy decoding to ensure reproducibility and evaluate the outputs by keyword matching since we observe that the outputs of Generative MLLMs are quite structured, showing their strong instruction-following ability.

The reason why we use different commands after the main question (e.g., ``\texttt{Answer left or right}'', ``\texttt{Choose from the two options}'', and ``\texttt{Give a short answer}'') is that we find the LLaVA-1.5 model sensitive to such command. For instance, we try ``\texttt{Answer on or under}'' and ``\texttt{Answer with under or on}'' for the On/Under subset in What’sUp Subset A, and the model accuracy is quite low. For Phi-3-V-3.8B and LLaMA-3-V-8B, we try these prompts and pick the one with the highest accuracy. This is one of their limitations that deserves future research. However, we aim to show that they can extract such information, so we use the best prompt to showcase its ability. 

The results are shown in Table~\ref{tab:response}. We observe that the accuracy of LLaVA-1.5-7B is increased on On/Under and Front/Behind subsets. However, the performance of LLaMA-3-V-8B is worsened. Overall, they still surpass CLIP.
\begin{table*}[t]
\begin{center}
\begin{tabular}{l|l}
\toprule
 Subset & Question \\
\midrule 
What'sUp Subset A\&B, Left/Right & \multicolumn{1}{p{10cm}}{\texttt{Is the (object 1) to the left of or to the right of the (object 2)? Answer left or right.}}\\
What'sUp Subset A, On/Under & \multicolumn{1}{p{10cm}}{\texttt{Is the (object 1) on or under the (object 2)? Choose from the two options.}}\\
What'sUp Subset B, Front/Behind & \multicolumn{1}{p{10cm}}{\texttt{Is the (object 1) in front of or behind the (object 2)? Answer front or behind.}}\\
What'sUp Subset A (four-way) & \multicolumn{1}{p{10cm}}{\texttt{Is the (object1) to the left of, to the right of, on, or under the (object2)? Choose from the four options.}}\\
What'sUp Subset B (four-way) & \multicolumn{1}{p{10cm}}{\texttt{Is the (object1) to the left of, to the right of, in front of, or behind the (object2)? Answer front, behind, left, or right.}}\\
COCO/GQA-spatial, One obj. & \multicolumn{1}{p{10cm}}{\texttt{Is the (object 1) on the (left/right/top/bottom) or on the (right/left/bottom/top)? Give a short answer.}}\\
COCO-spatial, Two obj. & \multicolumn{1}{p{10cm}}{\texttt{Is the (object 1) (to the left of/to the right of/above/below) a (object 2) or (to the right of/to the left of/below/above) a (object 2)? Give a short answer.}}\\
GQA-spatial, Two obj. & \multicolumn{1}{p{10cm}}{\texttt{Is the (object 1) to the (left/right/front/behind) of a (object 2) or to the (right/left/behind/front) of a (object 2)? Give a short answer.}}\\
\bottomrule
\end{tabular}
\caption{Question formats for different subsets for LLaVA-1.5-7B. }
\label{tab:prompts}
\end{center}
\end{table*}
\begin{table*}[t]
\begin{center}
\begin{tabular}{l|l}
\toprule
  Subset & Question \\
\midrule 
 What'sUp Subset A\&B, Left/Right & \multicolumn{1}{p{10cm}}{\texttt{Is the (object 1) to the left of or to the right of the (object 2)?}}\\
 What'sUp Subset A, On/Under & \multicolumn{1}{p{10cm}}{\texttt{Is the (object 1) at the bottom of the (object2) or at the top of the (object2)?}}\\
 What'sUp Subset B, Front/Behind & \multicolumn{1}{p{10cm}}{\texttt{Is the (object 1) in the back of the (object2) or in the front of the (object2)?}}\\
\bottomrule
\end{tabular}
\caption{Question formats for evaluating LLaVA-1.5-7B-VLM2Vec-LoRA.}
\label{tab:vlm2vec_prompts}
\end{center}
\end{table*}
\begin{table*}[t]
\begin{center}
\begin{tabular}{lcccccccc}
\toprule
& \multicolumn{4}{c}{What'sUp Subset A} & \multicolumn{4}{c}{What'sUp Subset B} \\
& \multicolumn{2}{c}{Left/Right} & \multicolumn{2}{c}{On/Under} & \multicolumn{2}{c}{Left/Right} & \multicolumn{2}{c}{Front/Behind}  \\
& Indiv. & Pairs & Indiv. & Pairs & Indiv. & Pairs & Indiv. & Pairs \\
\midrule
CLIP-ViT-L/14-336px & 49.0 & 1.9 & 61.7 & 23.3 & 54.9 & 10.8& 51.5& 7.8\\ 
\noalign{\vskip 0.3mm} \cdashline{1-9} \noalign{\vskip 0.3mm}
LLaVA-1.5-7B & 99.0& 98.1 & 80.1 & 60.2 & \textbf{100}& \textbf{100} & \textbf{98.5}& \textbf{97.1}\\ 
Phi-3-V-3.8B & \textbf{100} &\textbf{100}&\textbf{85.4}&\textbf{70.9}&\textbf{100}& \textbf{100} &56.9 &13.7\\
LLaMA-3-V-8B & 90.3 &80.6 &57.8&20.4 & 71.1 & 46.1 &69.1& 41.2 \\
\bottomrule
\end{tabular}
\caption{Results of CLIP-ViT-L/14-336px and Generative MLLMs on four subsets in What'sUp using standard response-based evaluation. The individual accuracy and pair accuracy are in percentage points. }
\label{tab:response}
\end{center}
\end{table*}
\subsection{Evaluating VLM2Vec}\label{appendix_eval_vlm2vec}
For evaluation, we use the same question template as for training (``\texttt{Represent the given image with the following question: \{Question\}}''). We list the questions used for VLM2Vec's evaluation in Table~\ref{tab:vlm2vec_prompts}. Similar to response-based evaluation for Generative MLLMs, we notice variance when using different questions. Here, we adopt the questions that lead to the best performance on the benchmarks.

In addition, we show that the benefit of using a question in the prompt generalizes beyond What'sUp. Here, we perform the same comparison as in Table~\ref{tab:vlm2vec} on MMVP and MMVP-VLM. We use the original question of the benchmark in the prompt. For MMVP-VLM which does not have questions, we manually add an MMVP-like question to each test case without altering content or tuning the prompt. We attached these questions to the updated supplementary material. We use the same prompt format as What'sUp (``\texttt{Represent the given image with the following question: \{Question\}}'' or ``\texttt{Represent the given image.}'' without any question). We observe similar results in Table~\ref{tab:vlm2vec-mmvp}.
\begin{table*}[t]
\begin{center}
\vspace{-10pt}
\begin{tabular}{lcc}
\toprule
& MMVP & MMVP-VLM \\
\midrule
CLIP-ViT-L/14-336px & 14.0&20.7\\ 
\noalign{\vskip 0.3mm} \cdashline{1-3} \noalign{\vskip 0.3mm}
LLaVA-1.5-7B-VLM2Vec-LoRA & \textbf{30.0}  & \textbf{37.8}\\
w/o Question in Prompt & 9.3&11.9\\
\midrule 
Random chance & 25.0 & 25.0  \\
\bottomrule
\end{tabular}
\vspace{-5pt}
\caption{The pair accuracy of CLIP-ViT-L/14-336px and LLaVA-1.5-converted models in percentage points on MMVP and MMVP-VLM. LLaVA-1.5-7B-VLM2Vec-LoRA continues to outperform CLIP. When there is no question in the prompt, its performance degenerates to the standard CLIP.}
\label{tab:vlm2vec-mmvp}
\end{center}
\end{table*}
\section{Supplementary Experimental Details and Results}
\subsection{Hyperparameters}\label{appendix_hyperparameters}
Our code for training standard CLIP, SigLIP, and EVA-CLIP is based on \url{https://github.com/mlfoundations/open_clip}~\citep{ilharco_gabriel_2021_5143773}. We finetune these models for five epochs with a learning rate of 5e-6 on the combination of converted LCS-558K plus converted DataMix-665K. We use 50 steps of warmup and AdamW optimizer with a cosine-annealing learning rate schedule. The batch size is 512, and we train the models on 4 GPUs. The training time is less than one day.

For the \texttt{[CLS]}-LLaVA-1.5-7B-LoRA and reproduced LLaVA-1.5-7B-LoRA in Table~\ref{tab:ablation_cls_llava}, we use the official LLaVA code in \url{https://github.com/haotian-liu/LLaVA} released under the Apache 2.0 license. The batch size, learning rate, and other training settings are the same as described in LLaVA-1.5 paper~\citep{liu2024improved}. 

For the experiments in Table~\ref{tab:ablation_pacl}, we start from an implementation of PACL~\citep{mukhoti2023open} in \url{https://github.com/NMS05/Patch-Aligned-Contrastive-Learning}. Since we only need to train the vision embedder and text embedder, we apply a larger batch size (4096) and train for 10 epochs on 8 GPUs on the combination of converted LCS-558K plus converted DataMix-665K. We use 0.1 as the fixed temperature, 1e-4 as the learning rate, and Adam as the optimizer. The training time is less than one day for all experiments. We adopt the data augmentation implemented in the codebase, except the \texttt{RandomHorizontalFlip}, which discourages models from learning about Left/Right spatial relations. When processing the captions, we follow the technique used in the codebase, which uses the original caption randomly with a probability of 0.5, and template+(a noun in the caption) otherwise. The templates are: ``\texttt{a picture of \{\}.}'',
            ``\texttt{itap of \{\}.}'',
            ``\texttt{a photograph of \{\}.}'',
            ``\texttt{this picture contains \{\}.}'',
            ``\texttt{a good photo of \{\}.}''. For experiments with a stronger text encoder, we do not apply this technique on DataMix-665K.

For LLaVA-1.5-7B-VLM2Vec-LoRA in Table~\ref{tab:vlm2vec}, we refer to the VLM2Vec code in \url{https://github.com/TIGER-AI-Lab/VLM2Vec}. We use rank=8 for LoRA, 256 for the batch size, 1024 for maximum input token length, and 0.02 for the temperature. We train the model for only 900 steps on 4 GPUs for 40 hours on the combination of LCS-558K and DataMix-665K, with a linear learning rate schedule, 100 warmup steps, and 2e-5 as the learning rate. Although we do not train the model on full data, the model performance is remarkable on What'sUp.
\subsection{Converting LLaVA-1.5's Training Data}\label{appendix_data}
We use LLaVA-1.5's training data for all finetuning experiments we include in the paper. The DataMix-665K is under CC BY 4.0 License, while the LCS-558K is under LAION/CC/SBU License for images and BSD 3-Clause "New" or "Revised" License for BLIP-generated captions. They are datasets of English conversations.

We check the frequency of appearance of the following keywords in DataMix-665K and LCS-558K: \texttt{on the left}, \texttt{on the right}, \texttt{to the left}, \texttt{to the right}, \texttt{at the left}, \texttt{at the right}. In DataMix-665K, there are 12957 instances with at least one of the key phrases, among which 12658 have a paired image. For captions (ground truth answers), this number is 13473 since an instance is paired with a multi-turn conversation. In LCS-558K, there are 560 such instances and captions since each instance has only one question and one answer.

In our experiments in Section~\ref{ab1_data} and Section~\ref{ab2_token}, LCS-558K was converted from image-text pair format to conversation format, so we revert this process by using ground truth answer as the caption. Since DataMix-665K is in a multi-turn conversation format, we randomly pick one answer as the caption in each epoch. In Section~\ref{ab3_lm_choice}, the new text encoder can encode long paragraphs, so we use the concatenation of all answers in the multi-turn conversation as the ground truth caption. In practice, we calculate the text embeddings of all possible captions using the text encoder of LLM2CLIP before training to save memory and time. In Section~\ref{ab4_vlm2vec}, we randomly choose one turn from the multi-turn conversation.
\subsection{Results of Unlocking Image Encoder}\label{appendix_unlock}
We try unlocking the image encoder during finetuning on the SigLIP-ViT-L/16-384px model. The results are in Table~\ref{tab:unlock_data}. Still, the individual accuracy remains low.
\label{appendix_ablation}
\begin{table*}[t]
\begin{center}
\begin{tabular}{lccccccccc}
\toprule
& \multicolumn{2}{c}{What'sUp Subset A}&\multicolumn{2}{c}{What'sUp Subset B}&\multicolumn{2}{c}{COCO-spatial} & \multicolumn{2}{c}{GQA-spatial} \\
 & Indiv. & Pairs & Indiv. & Pairs & One-obj. & Two-obj. & One-obj. & Two-obj. \\
\midrule 
SigLIP-ViT-L/16-384px &50.0&1.9&51.5&5.9&48.7&50.2&51.2&47.0\\
+ finetuning (ft) & 50.5&2.9&51.5&5.9&48.7&57.7&50.5&48.1\\
+ ft + hard neg. & 50.0&3.9&47.1&2.0&52.3&47.0&51.8&52.7\\
\midrule 
Random chance & 50.0 & 25.0  & 50.0 & 25.0  & 50.0 & 50.0 & 50.0 & 50.0 \\
\bottomrule
\end{tabular}
\caption{Results of SigLIP-ViT-L/16-384px focusing on the Left/Right subsets of What'sUp, COCO-spatial, and GQA-spatial benchmark with unlocked image encoder, after finetuning on LLaVA-1.5's training data with or without hard negative captions. The accuracy remains low on all benchmarks.}
\label{tab:unlock_data}
\end{center}
\end{table*}
\subsection{Evaluating and Finetuning XVLM}\label{appendix_xvlm}
\begin{table*}[t]
\begin{center}
\begin{tabular}{lccccccccccc}
\toprule
& What'sUp A&What'sUp B&\multicolumn{2}{c}{COCO-spatial} & \multicolumn{2}{c}{GQA-spatial} \\
 & & & One-obj. & Two-obj. & One-obj. & Two-obj. \\
\midrule 
XVLM-16M &\textbf{50.0}&32.8&65.4&64.6&\textbf{63.2}&\textbf{53.3}\\
+ finetuning& 46.4 &\textbf{34.6}&\textbf{66.8}&\textbf{65.2}&61.3&51.2\\
\midrule 
Random chance & 25.0  & 25.0 & 50.0 & 50.0 & 50.0 & 50.0 \\
\bottomrule
\end{tabular}
\caption{Individual accuracy of XVLM-16M on the Left/Right subsets of What'sUp, COCO-spatial, and GQA-spatial benchmark on LLaVA-1.5's training data. LLaVA-1.5's training data does not help improve XVLM-16M notably.}
\label{tab:xvlm_data}
\end{center}
\end{table*}
Observing the similar failure of the data-informed attempt, previous work concluded that even with relevant, high-quality data and hard negatives, denser supervision is likely required to let the model learn the basic spatial relations~\citep{kamath2023s}, as in XVLM~\citep{zeng2021multi}, a VLM with supervision at the bounding-box level. We attach XVLM-16M's performance in Table~\ref{tab:xvlm_data}. We find that Generative MLLMs still beat XVLM-16M, while they do not incorporate downstream task-related inductive bias or denser supervision. 

We explore finetuning XVLM on LLaVA-1.5's training data based on their official code (\url{https://github.com/zengyan-97/X-VLM}), but no improvement is observed in the results (the last row in Table~\ref{tab:xvlm_data}). The image encoder is locked during finetuning. We use both contrastive learning loss and image-text matching loss to finetune the XVLM-16M model for five epochs with a learning rate of 1e-5 and a weight decay rate of 0.01. We use 10\% steps of warmup and AdamW optimizer with a lambda learning rate schedule. The batch size is 128, and we train the model on 4 GPUs. The evaluation is performed through the image-text matching score. 
\subsection{Implementation Details of PACL and SPARC}\label{appendix_sparc}
For PACL, the vision embedder applied on CLIP-ViT-L/14-336px is the sum of a one-layer linear projection and a two-layer nonlinear projection with GELU as the activation function. The input of the vision embedder is 576 1024-dimensional patch tokens after \texttt{LayerNorm} and \texttt{Dropout} (with probability = 0.1), and the output dimension is 768 for 576 tokens. The text embedder accepts one 768-dimensional text token and applies one-layer linear projection on it. All output embeddings are L2-normalized. The embedders used in SPARC experiments share the same model structures as in PACL. For experiments with the text encoder from LLM2CLIP, the input dimension of the text embedder is 1280 instead of 768, with other settings unchanged. 

For RoPE, we refer to the implementation in the codebase for LLaMA in \url{https://github.com/huggingface/transformers}. It is applied before \texttt{LayerNorm}. Compared with learned or sinusoidal position embeddings, it maintains the relative positions of tokens. We choose to use it since it is applied in language models of Generative MLLMs listed in Section~\ref{sec:compare}.

For SPARC, we follow the pseudocode in Appendix C of \citet{bica2024improving}. Specifically, the output of the vision embedder is 576 768-dimensional projected patch tokens, and the output of the text embedder is 77 768-dimensional projected text tokens (with padding). After multiplying them, we get a similarity matrix of size $77\times 576$. Following the SPARC paper, we first apply min-max normalization to the matrix and then sparsify it by zeroing out all matrix entries below the threshold $1/576$. We normalize the rows of the similarity matrix, multiply it with the patch tokens, and obtain 77 grouped visual tokens. The global representation of the image is the mean of these grouped visual tokens after L2-normalization, and we get the global representation of the text similarly. During inference, we calculate the cosine similarity between the two global representations. When training, we use the global representations for the standard contrastive loss and apply a local contrastive loss to contrast the 77 grouped visual tokens and 77 text tokens within each sample. In this way, we align the patch tokens to individual concepts represented by text tokens. Unlike the original implementation, we do not use a learnable temperature for contrastive losses.
\subsection{When Generative MLLMs are worse than CLIP}
We also observe that in some cases, MLLMs have worse performance than CLIP (See Table~\ref{tab:failure}). On EqBen-mini~\citep{wang2023equivariant}, their performance is close. On COCOCounterfactuals~\citep{le2024coco}, we notice that CLIP embeddings are involved in the construction of the benchmark as a metric, which could affect the comparison. 

This phenomenon is also discussed in previous literature~\citep{zhang2024visually, geigle2024african}, where they find that training on web-crawled data teaches CLIP many rare concepts, while Generative MLLMs are not sufficiently exposed to such data for image-text alignment. 
\begin{table*}[t]
\begin{center}
\begin{tabular}{lcccccccc}
\toprule
& EqBen-mini &COCOCounterfactuals  \\
\midrule
CLIP-ViT-L/14-336px & \textbf{40.0} &\textbf{87.7} \\ 
LLaVA-1.5-7B & 32.9 & 57.9\\ 
\bottomrule
\end{tabular}
\caption{The pair accuracy of CLIP-ViT-L/14-336px and LLaVA-1.5-7B in percentage points. }
\label{tab:failure}
\end{center}
\vspace{-15pt}
\end{table*}
\subsection{Alternative Decoding for Generative MLLMs}\label{appendix_m3id}
\setlength{\tabcolsep}{2pt}
\begin{table*}[htbp]
\begin{center}
\begin{tabular}{lccc}
\toprule
& Indiv. & Pairs\\
\midrule 
LLaVA-1.5-7B &  61.7& 25.3 \\
+ M3ID~\citep{favero2024multi} & \textbf{64.3} &\textbf{31.3}\\
LLaVA-1.5-13B + I-MoF~\citep{tong2024eyes} & -- & \textbf{31.3} \\ 
\midrule 
Random chance & 50.0  & 25.0  \\
\bottomrule
\end{tabular}
\caption{Results of LLaVA-1.5-7B with M3ID ($\alpha=0.6, \lambda=0.15$) using response-based evaluation on MMVP benchmark with the original results. M3ID encourages LLaVA-1.5-7B to attend more to the visual input, achieving performance on par with using interleaved CLIP and DINO features.}
\label{tab:decoding}
\end{center}
\vspace{-15pt}
\end{table*}
We apply Multi-Modal Mutual-Information Decoding (M3ID)~\citep{favero2024multi} on LLaVA-1.5 for response-based evaluation. For token in each decoding step $t$, M3ID computes the output probability with the image and without any input image, denoted as $\mathbf{l_c}$ and $\mathbf{l_u}$ respectively. The latter corresponds to the language priors of the answer to the given question. Then a correction term ($\mathbf{l_c}-\mathbf{l_u}$) is added to $\mathbf{l_c}$ with weight $\frac{1-\exp(-\lambda t)}{\exp(-\lambda t)}$ if the model is not highly confident with the token in step $t$ ($\max_k (l_c)_k < \log \alpha$ where $\alpha, \lambda$ are pre-defined hyperparameters). This correction prevents the VLM from omitting the visual input and relying on the language priors.

In Table~\ref{tab:decoding}, this method achieves gain (+6\%) relative to the baseline LLaVA-1.5-7B on MMVP. We note that this is on par with I-MoF with interleaved CLIP and DINO features)~\citep{tong2024eyes}. This result suggests that LLaVA-1.5 did not attend to the visual input enough and thus might miss the key information for answering the query. A similar finding was described through the interpretability perspective on attention weights in \citet{stan2024lvlm}.

\end{document}